\documentclass[sigconf]{acmart}
\acmSubmissionID{577}

\usepackage{wrapfig}

\usepackage{enumitem}

\usepackage[ruled]{algorithm2e} 

\SetAlFnt{\small}
\SetAlCapFnt{\small}
\SetAlCapNameFnt{\small}
\SetAlCapHSkip{0pt}

\acmJournal{TOG}

\copyrightyear{2023}
\acmYear{2023}
\setcopyright{acmlicensed}\acmConference[SIGGRAPH '23 Conference Proceedings]{Special Interest Group on Computer Graphics and Interactive Techniques Conference Conference Proceedings}{August 6--10, 2023}{Los Angeles, CA, USA}
\acmBooktitle{Special Interest Group on Computer Graphics and Interactive Techniques Conference Conference Proceedings (SIGGRAPH '23 Conference Proceedings), August 6--10, 2023, Los Angeles, CA, USA}
\acmPrice{15.00}
\acmDOI{10.1145/3588432.3591535}
\acmISBN{979-8-4007-0159-7/23/08}

\newcommand{\method}{PhotoMat}
\newcommand{\mlp}{\textbf{M}}
\newcommand{\gen}{\textbf{G}}
\newcommand{\decoder}{\textbf{E}}
\newcommand{\loss}{\mathcal{L}_E}

\newcommand{\relit}{{\textbf{I}_{\tiny nr}}}
\newcommand{\real}{{\textbf{I}_{\tiny real}}}
\newcommand{\realcrop}{{\textbf{I}_{\tiny crop}}}
\newcommand{\rendered}{{\textbf{I}_{\tiny ar}}}

\newcommand{\xz}[1]{\textcolor{black}{{#1}}}

\begin{document}

\title{PhotoMat: A Material Generator Learned from Single Flash Photos}

\author{Xilong Zhou}
\email{zhouxilong199213@tamu.edu}
\affiliation{%
 \institution{Texas A\&M University}
 \institution{Adobe Research}
 \city{College Station}
 \country{USA}}

\author{Milo\v s Ha\v san}
\email{milos.hasan@gmail.com}
\affiliation{%
 \institution{Adobe Research}
  \city{San Jose} 
 \country{USA}
}

\author{Valentin Deschaintre}
\email{deschain@adobe.com}
\affiliation{%
 \institution{Adobe Research}
 \city{London} 
 \country{UK}
}

\author{Paul Guerrero}
\email{guerrero@adobe.com}
\affiliation{%
 \institution{Adobe Research}
 \city{London} 
 \country{UK}
}

\author{Yannick Hold-Geoffroy}
\email{holdgeof@adobe.com}
\affiliation{%
 \institution{Adobe Research}
 \city{San Jose}  
 \country{USA}
}

\author{Kalyan Sunkavalli}
\email{sunkaval@adobe.com}
\affiliation{%
 \institution{Adobe Research}
 \city{San Jose}  
 \country{USA}
}

\author{Nima Khademi Kalantari}
\email{nimak@tamu.edu}
\affiliation{%
 \institution{Texas A\&M University}
 \city{College Station}
 \country{USA}}

\renewcommand\shortauthors{Zhou et al.}

\begin{teaserfigure}
    \centering
	\includegraphics[width=\linewidth]{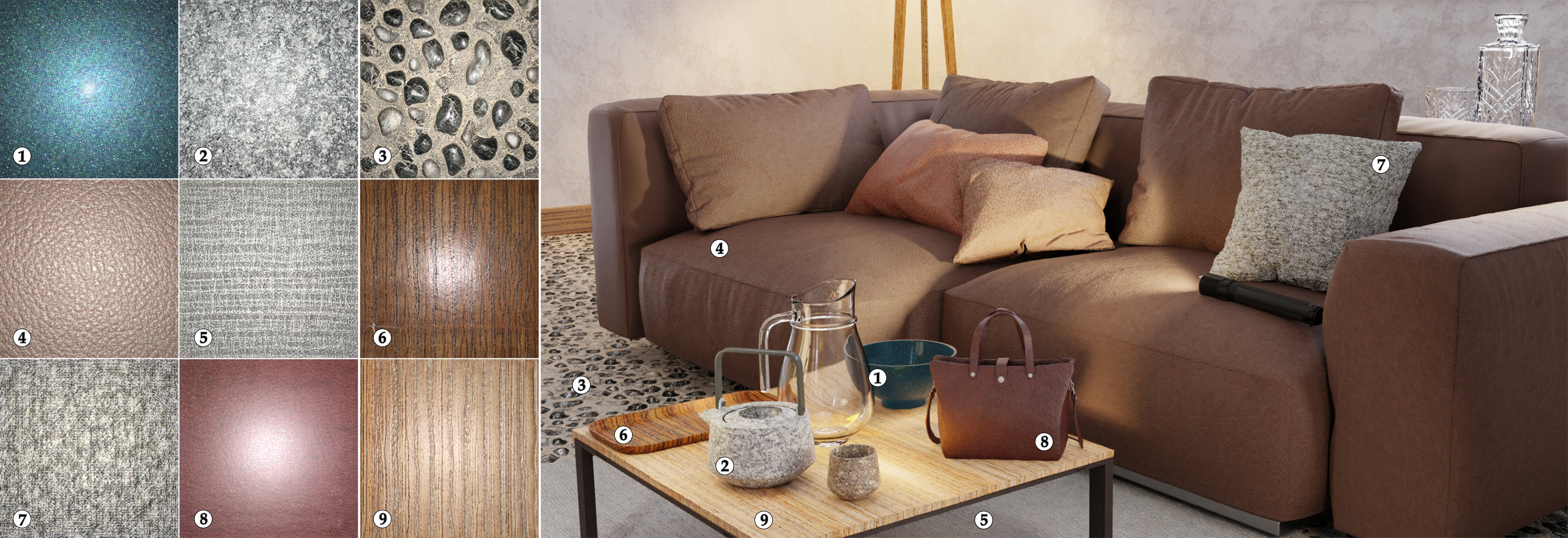}
	\caption{We present a material generative model trained exclusively on real cell-phone flash photographs. Our model produces plausible photorealistic materials (left), which can readily be applied to 3D scenes (right). All 18 materials (except glass) are from our generator. Geometry by Flavio Della Tommasa (CC-BY).} 
	\label{fig:teaser}
\end{teaserfigure}
 
\begin{abstract}
Authoring high-quality digital materials is key to realism in 3D rendering.
Previous generative models for materials have been trained exclusively on synthetic data; such data is limited in availability and has a visual gap to real materials. 
%
We circumvent this limitation by proposing \method{}: the first material generator trained exclusively on \emph{real photos} of material samples captured using a cell phone camera with flash. 
Supervision on individual material maps is not available in this setting. 
Instead, we train a generator for a neural material representation that is rendered with a learned relighting module to create arbitrarily lit RGB images; these are compared against real photos using a discriminator. 
We train PhotoMat with a new dataset of 12,000 material photos captured with handheld phone cameras under flash lighting.
We demonstrate that our generated materials have better visual quality than previous material generators trained on synthetic data.
Moreover, we can fit analytical material models to closely match these generated neural materials, thus allowing for further editing and use in 3D rendering.



%

\end{abstract}

%
%


%
%


\maketitle

\section{Introduction}

Materials with accurate spatially-varying reflectance properties are among the most important factors determining the realism of rendered scenes. These materials are often represented as spatially varying parameters of analytic BRDFs, commonly termed SVBRDFs. 
Unfortunately, creating high-quality materials is challenging. Authoring materials from scratch usually requires significant time and experience with specialized tools like Substance Designer~\cite{SubstanceDes}. An alternative is to capture material reflectance using specialized hardware such as spherical gantries or domes. This also has found limited use due to the cost of the acquisition process. 

In recent years, researchers have proposed data-driven methods to either capture or generate material reflectance without the need for specialized skills or hardware. For lightweight capture, many convolutional image-to-image translation models to map one or few RGB photos of materials to output material maps have been proposed~\cite{Deschaintre2018, Gao2019,Deschaintre2020,Guo2021,Martin2022}. Other works have proposed methods for unconditional or conditional generation of materials~\cite{Guo2020,zhou2022tilegen}. All these methods (both capture and generation) require explicit supervision on individual material maps, and as a result, are trained on synthetic material datasets.
Because of the cost of authoring digital materials, these datasets are limited in number and diversity; for example, most of these methods are trained on the \citet{Deschaintre2018} material dataset that has only 1850 materials derived from 155 procedural materials in it.
Moreover, even though they are carefully designed, these synthetic materials still have a substantial visual gap to real-world materials.
As a consequence, the methods synthesize results that inherit the synthetic data distribution, giving them a certain digital look. 

To address this challenge, we ask the question: is it possible to train a \xz{realistic material generator without using synthetic maps}?
This is a chicken-and-egg problem: to train a material generator we need a large dataset of real reflectance maps, while to produce such a dataset, we need methods to acquire realistic materials from one or few images. 
Our key idea to solve this problem is to perform both generation and acquisition within a \emph{single framework}. In contrast to existing work, we present \method{}, a material generation model trained on a large dataset of easy to capture flash photographs, with known light source (flash) location. 
Our key contribution is a specifically designed system that ensures that \method{} learns real-world reflectance properties without directly observing ground truth material maps.

\method{} uses a high-dimensional ``neural material'' map generator (based on StyleGAN2 \cite{StyleGAN2}) that produces relightable per-pixel features that encode the appearance of that point.
These neural features are then fed into a per-pixel neural relighting multi-layer perceptron (MLP) to render images under a conditional light source (flash) location.
A discriminator, also conditioned on the flash light location, takes these renderings as well as cropped samples from the real dataset as inputs. Through training, we learn both the generator that can produce a powerful implicit neural representation of reflectance properties, as well as a relighting module that can simulate the rendering process given a chosen light location. Together, these two modules produce rendered images which follow the distribution of the real flash images. 

We use our relightable neural material representation to explicitly estimate analytical material maps which, when rendered, closely match the relightable material.
To achieve this, we train a material map estimator that takes the GAN-generated neural material as input and produces per-pixel material parameters that are then fed into an analytic differentiable renderer. The loss between the analytically and neurally rendered materials is back-propagated to train the material map estimator. We demonstrate that \method{} can learn a generic material generator for standard microfacet-based analytical material models, as well as specialized generators for specific materials (e.g. coated BRDFs for car paints).

Finally, to gather a large scale dataset of flash images with known light location, we propose a simple, but effective, data collection mechanism that eliminates the need for camera calibration. Each image is collected using a handheld phone camera with flash light on under weak ambient environment so that flash light dominates the material surface. We then apply a simple light detector to detect flash light position of each image. 
We use this casual capture process to collect and process a large dataset of 12,000 real photos. 
In summary, this paper makes the following contributions: 

\begin{itemize}
\item PhotoMat, the first material generative model trained exclusively on real photos without relying on a specific analytic BRDF model or an existing SVBRDF dataset.

\item A neural material representation that can be decoded into analytic SVBRDF parameters with no material map supervision.

\item A data collection strategy (using a handheld phone with flash) can be easily scaled to large material datasets. We collect 12,000 material flash photos that will be publicly released. 
\item Material generation of high-quality SVBRDFs, across several material categories, that can be used in practical 3D rendering.
\end{itemize}
\section{Related Work}

As an essential part of the rendering pipeline, material creation and editing have received significant attention from both researchers and industry practitioners.
In the computer graphics industry, material creation is typically done using procedural node graph editing software~\cite{SubstanceDes}, but material acquisition and generative models are growing in importance.

\paragraph{Material acquisition} Traditional acquisition approaches rely on extensively sampling both light and view directions using a gonioreflectometer~\cite{Matusik2003, Guarnera2016}. Recent methods have relied on data-driven approaches to retrieve material properties from a single image or a few images, with either flash or natural lighting. Methods using different architectures have been proposed~\cite{Deschaintre2018, Li2017, Li2018, Deschaintre2019, Gao2019, Deschaintre2020, Guo2021, Zhou2021, zhou2022look, Martin2022}. All of these approaches rely on supervised training with synthetic data. Zhou et al.~\shortcite{Zhou2021} use an adversarial loss for training and complement the synthetic data with a small real dataset, and Henzler et al.~\shortcite{Henzler2021} rely on a small dataset of captured flash images for pre-training, but require a fine-tuning step to reproduce the input image appearance. As opposed to our goal, these methods target material acquisition from a flash picture; neither of these approaches can generate new materials. 
In fact, many of the methods above ultimately rely on the \emph{same} datasets, Substance Source and Substance Share, and the derived dataset by ~\citet{Deschaintre2018}, highlighting the need for material research to move beyond limited synthetic data.


\paragraph{Material generation}
MaterialGAN~\cite{Guo2020} trains an unconditional generative model for SVBRDFs. Their main goal is to optimize in the generator latent space to match the appearance of a real material given a few target pictures, thus the quality of the generated materials is only a secondary goal, since they are not used directly.
TileGen~\cite{zhou2022tilegen} extends MaterialGAN, focusing on a per class approach. This approach improves the architecture to enable tileable SVBRDF generation that can be conditioned on the input patterns. Hu et al. \shortcite{hu2022control} use a network similar to TileGen~\cite{zhou2022tilegen} as a prior for material appearance transfer.

An alternative material representation heavily used in industry is procedural node graphs. Their creation typically relies on professional software and requires specialized skills. To simplify their creation, Guerrero et al.~\shortcite{guerrero2022matformer} proposed an unconditional procedural graph generator, leveraging transformers to generate tokenized nodes, edges and parameters of material graphs. Recently, EG3D~\cite{Chan2021} proposed a GAN for 3D face generation supervised only with 2D images; this problem is related to ours, since it generates an asset type that is not directly observed in 2D photos but required in 3D scene authoring. We base our generative models on the StyleGAN 2 architecture~\cite{StyleGAN2}, which is widely used for image synthesis in many domains. We modify the architecture to produce per-pixel neural feature maps, which are turned into final RGB values using a neural relighting MLP module.


\paragraph{Neural materials}
\method{} produces an intermediate neural material representation that can be rendered into arbitrarily lit photographs through an additional MLP-based relighting module. This is akin to neural materials that combine feature textures with an MLP decoder and have been directly used for rendering \cite{Rainer2020Unified,kuznetsov2021neumip,NeuralBSDF}. These works however require a large amount of data sampling to fit a specific material, and do not generate new materials. 
\section{The \method{} Method}

Our goal is to train a generative model producing material maps (such as albedo, roughness, normals, etc.) using a real dataset of flash material photos. The key problem is that the material maps are hidden variables that are not directly observed in the real RGB photos, so previous techniques supervised with ground-truth maps~\cite{Guo2020, zhou2022tilegen} cannot be applied here.

\subsection{Key idea}

Our idea is to split the problem into two parts. 
First, we train a conditional \emph{relightable} GAN for material images in the RGB domain: conditional on a desired flash highlight location, the generator is able to produce a rendered material image with the highlight in the requested place. This is done by generating an implicit neural material representation and using the conditional information with a neural relighting module in the last step to obtain a relit RGB image. Second, this relightability property lets us generate the same material under many known lighting conditions. For common BRDF models this is sufficient to decode explicit material parameters per texel that fit the appearance of the texel under the specified lighting conditions. We train another network to do this decoding from an implicit neural material to an explicit analytical material.

\begin{wrapfigure}{r}{0.5\columnwidth}
    \centering
    \includegraphics[width=0.5\columnwidth]{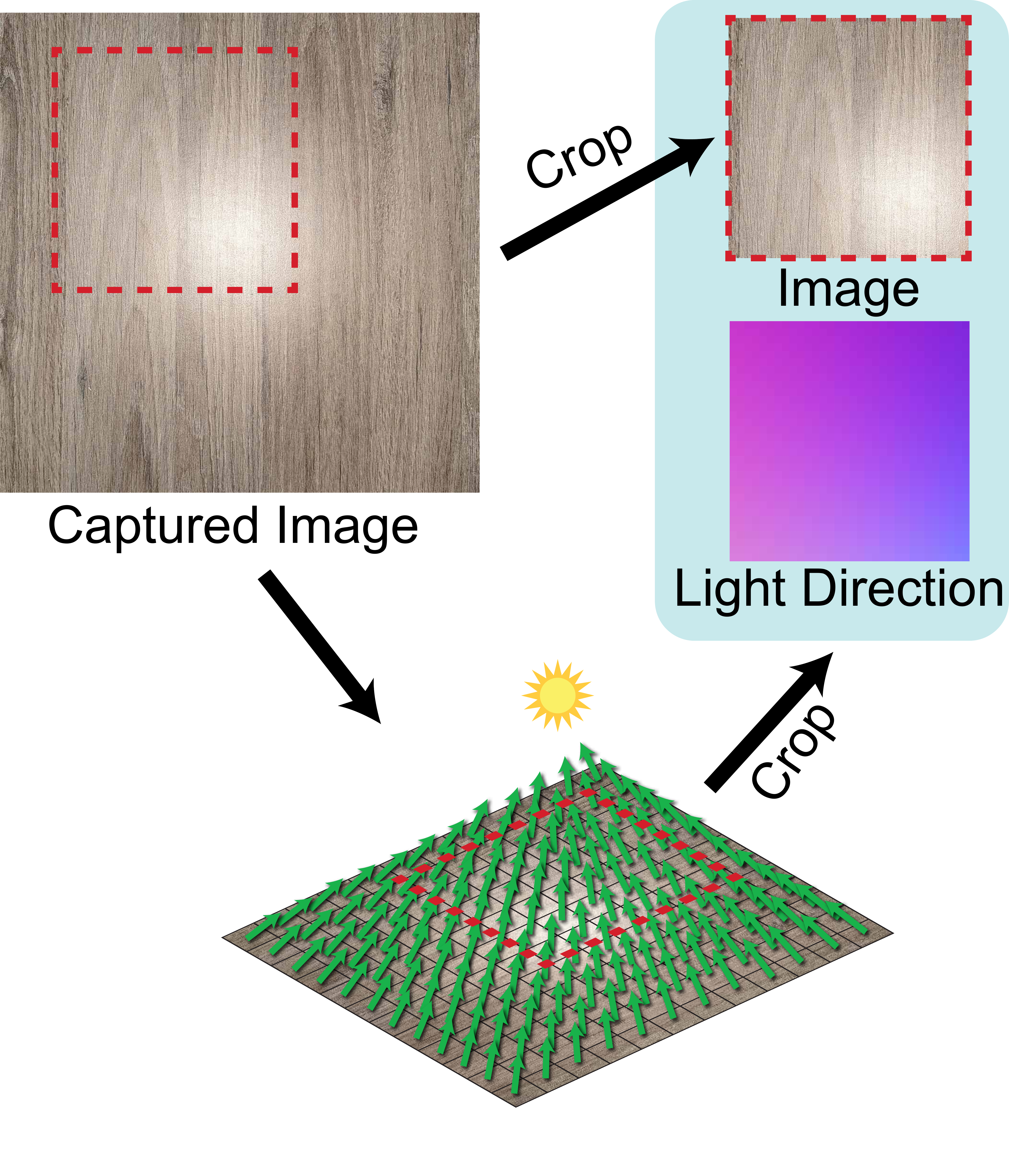}
\caption{We randomly crop photos to $\frac{1}{2}$ size to create crops with random highlight locations; each texel of each crop has a known light direction.}
\label{fig:dataset_crop}
\end{wrapfigure}

Training this method requires a dataset with a variety of flash highlight locations. Fortunately, we do not have to explicitly capture the materials with separate varying light sources. We can use simple cropping of the original photos to obtain images with a variety of highlight locations. The uncropped square photos have flash highlights approximately in the center (we detect the exact location); the light (and camera) direction can therefore be computed for each pixel of each crop (Fig. \ref{fig:dataset_crop}). 

Note that the photos, taken with a cell phone camera with flash, can be treated as having a \emph{collocated} camera and point light, where the slight distance between the physical flash and camera is negligible. Therefore, whenever we refer to the \emph{light location}, we imply this to be the same as the \emph{camera location}. Similarly, \emph{highlight location} means the point on the material sample directly below the light (camera). We can use all of these terms interchangeably, as they are trivially convertible to each other. Thus, by cropping photos in which the highlight is centered, we can achieve cropped material samples lit with a variety of highlight locations (and therefore a variety of light/camera locations with respect to the image center). 

\begin{figure*}
    \centering
    \includegraphics[width=1\linewidth]{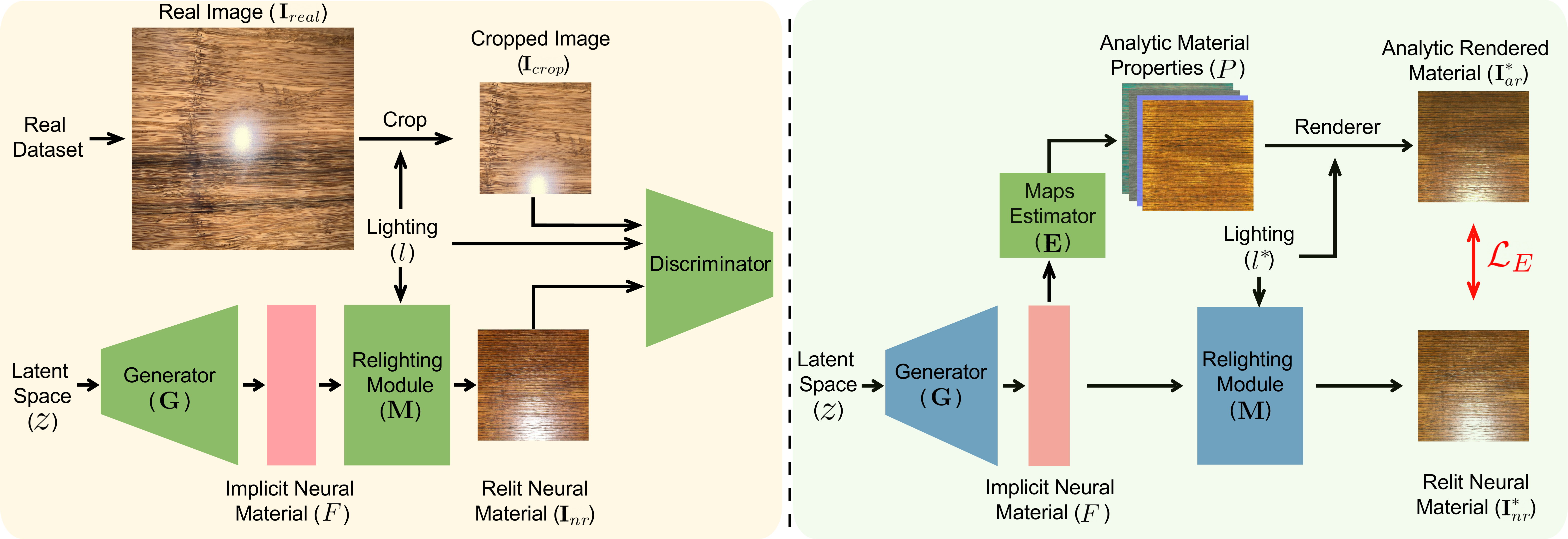}
\caption{A visual illustration of our method. {\bf Left:} In our relightable generator, a latent code $z$ feeds into the feature generator $\gen$ to produce an implicit neural material representation $F$, where every texel holds a 32-dimensional feature vector; optionally, $F$ can be tileable. Next, $F$ is fed to the conditional relighting module $\mlp$ to produce a relit neural material $\relit$ under highlight location $l$. During training, the discriminator takes either the relit neural material $\relit$ or a cropped real image $\realcrop$, and attempts to distinguish them. {\bf Right:} In the next step, we freeze the pretrained $\gen$ and $\mlp$, and train a convolutional decoder $\decoder$ to turn the neural material representation $F$ into per-texel analytic reflectance properties $P$. The decoder is trained to minimize the loss $\loss$ between re-rendered analytic materials $\rendered^*$ and relit neural materials $\relit^*$. Once all networks are trained, we can generate new materials by randomly sampling $z$, producing implicit neural material representation $F = \gen(z)$ and decoding them into analytic material properties $P = \decoder(F)$.}
\label{fig:architecture}
\vspace{-0.1in}
\end{figure*}

In summary, the \method{} solution consists of two steps: first, we propose a relightable generator to generate material images under conditional light source (highlight) locations (shown in the left part of the Fig.~\ref{fig:architecture}), trained on real images only. Second, we use a BRDF parameter estimator to decode analytical material reflectance parameters from the implicit neural material representation produced by the generator (shown in the right part of the Fig.~\ref{fig:architecture}). In the rest of this section, we provide more details.

\subsection{Relightable generator}

Generative adversarial networks (GANs) ~\cite{Goodfellow2014} have been successful generative models in many areas (images, video, audio). We build on recent work on material generative models, TileGen~\cite{zhou2022tilegen}, which is built upon a tileable version of StyleGAN2~\cite{StyleGAN2} to produce plausible tileable and controllable materials conditional on mask patterns. However, like all previous material generative models, TileGen is trained on synthetic data. Our architecture is also partially inspired by the EG3D human face generator conditional on camera pose~\cite{Chan2021}, where a StyleGAN-based generator produces a 3D feature field (tri-plane) of a human face, which is then passed through a neural renderer producing a 2D image of a face from a specific view.

The architecture of our relightable generator consists of a feature generator $\gen$ and a conditional neural relighting module $\mlp$. 
Given a latent code $z$, the feature generator $\gen(z)$ produces an implicit material representation $F$ with channel number $C=32$ and the same resolution as the final generated material. Next, $F$ is fed into the relighting MLP module $\mlp(F, l)$ conditioned on the highlight location $l$ to produce a ``neurally'' relit RGB image $\relit$:
\begin{equation} \label{eq:gan_syn}
	\relit(z, l) = \mlp(\gen(z), l) = \mlp(F, l).
\end{equation}
Note that $F$ can be seen as a neural representation for material reflectance. Similarly, $\mlp$ can be seen as a neural analogy to a classical local illumination renderer, where each texel of the material surface is independently lit based on the local light direction. We therefore convert the highlight location $l$ into per-texel light direction $\omega$ and use an MLP to implement $\mlp$ with local light direction $\omega$ and local feature vector $F$ for each individual texel as input. 
We show in Fig. \ref{fig:256_relight} several examples of relit images produced by our generator.

To train this combination of neural material generator $\gen$ and neural relighter $\mlp$ we use an adversarial loss with real flash images as training data. For each training step, we fetch a real example $\real$ from the real dataset, and crop a random region out of the original real materials to obtain the ``relit real material'' $\realcrop$. For example, for training a $256^2$ generator, we crop a random $256^2$ region out of a real photo resized to $512^2$; details about resolution are discussed in Sec. \ref{sec:impl}. We convert the highlight location in $\real$ to the corresponding highlight location in $\realcrop$, based on the crop boundaries, to account for this cropping operation. 

To train the discriminator in the image domain with our generated features, we train the relighting module to simulate a collocated light and view rendering process given generated features and a light direction. The discriminator is conditioned on $l$ and takes $\realcrop$ and the neural rendering $\relit$ for training. Once trained, our system can produce implicit neural material features that can be relit, using our relighting module, to produce renderings which follow the distribution of real flash photographs. We now describe how we use the generated features of our trained network, which contain all information to relight the represented material, as the input to a map estimator network. 

\xz{ In PhotoMat, $\gen$ and $\mlp$ are trained in a way where the inherent ambiguities between lighting and SVBRDF can be avoided. The $\gen$ and $\mlp$ models are trained with a GAN loss and varying lighting configurations, preventing baking highlights in neural materials $F$. Indeed, if the specular highlights were baked $F$, the discriminator would easily detect it due to the mismatch between the light condition and the observed highlight location.}

\subsection{Material map estimator}

Once our neural material generator $\gen$ and relighter $\mlp$ are trained, we aim to complete an end-to-end system to estimate analytical reflectance parameters $P$ from the generated implicit representation. 
We train a material estimator (decoder) $\decoder$, which takes the generated neural material $F$ as input and outputs the corresponding per-texel reflectance parameters $P$. The goal is to produce $P$ that, when rendered under different lighting conditions using an analytical differentiable renderer $\mathcal{R}$ to produce images $\rendered(P)$, approximates the image produced by the neural relighting module, $\relit(F)$.  The loss $\loss$ between analytically rendered materials and neurally rendered materials is backpropagated to train $\decoder$. Each training iteration samples a random latent code $z$ and a random lighting $l$, and our loss function is described as below:

\begin{equation}
	\label{eq:maps_estimator}
    \loss = \mathcal{L}(\rendered, \relit(z, l)), 
\end{equation}
where the analytically rendered image $\rendered$ is defined as:
\begin{equation}
	\label{eq:render}
    \rendered = \mathcal{R}(p, l) = \mathcal{R}(\decoder(\gen(z)), l).
\end{equation}
Here $\mathcal{R}$ represents the differentiable analytic renderer for the appropriate material model (e.g. a standard GGX-based microfacet model \cite{Walter2007}, or a more specialized model). Note that $l$ represents a randomly sampled lighting on a plane used for both the analytic and neural path, forcing consistency between $\rendered$ and $\relit$ for a given highlight location. For the loss function $\mathcal{L}$, we use the distance between Gram matrices of VGG layers~\cite{Gatys2015} combined with L1 loss, similar to TileGen~\cite{zhou2022tilegen}. The weight of gram matrix term is set as $1.0$ and L1 term is $0.1$. \xz{Note that $\decoder$ takes the lighting-independent neural materials $F$ as input and is forced to produce valid analytic BRDF parameter maps across different lighting configurations, avoiding baked lighting artifacts.}

\begin{figure}
    \centering
    \includegraphics[width=1\linewidth]{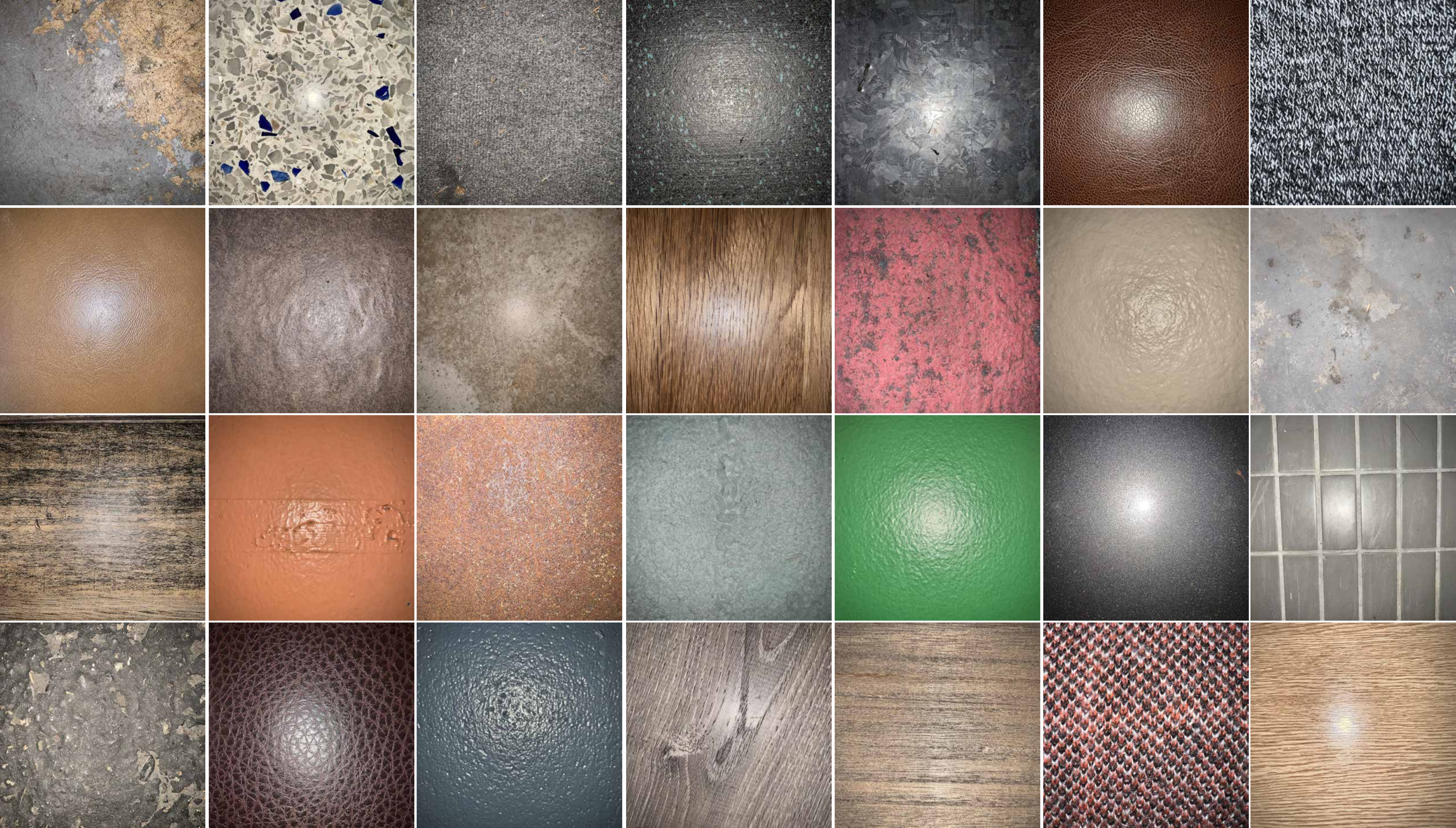}
\caption{Example flash photos from our in-the-wild dataset. We roughly center the highlight and avoid strong ambient lighting. Our pipeline is robust against small imperfections and does not require image calibration.}
\vspace{-0.2in}
\label{fig:real_dataset}
\end{figure}

\begin{figure}
    \centering
    \includegraphics[width=0.9\linewidth]{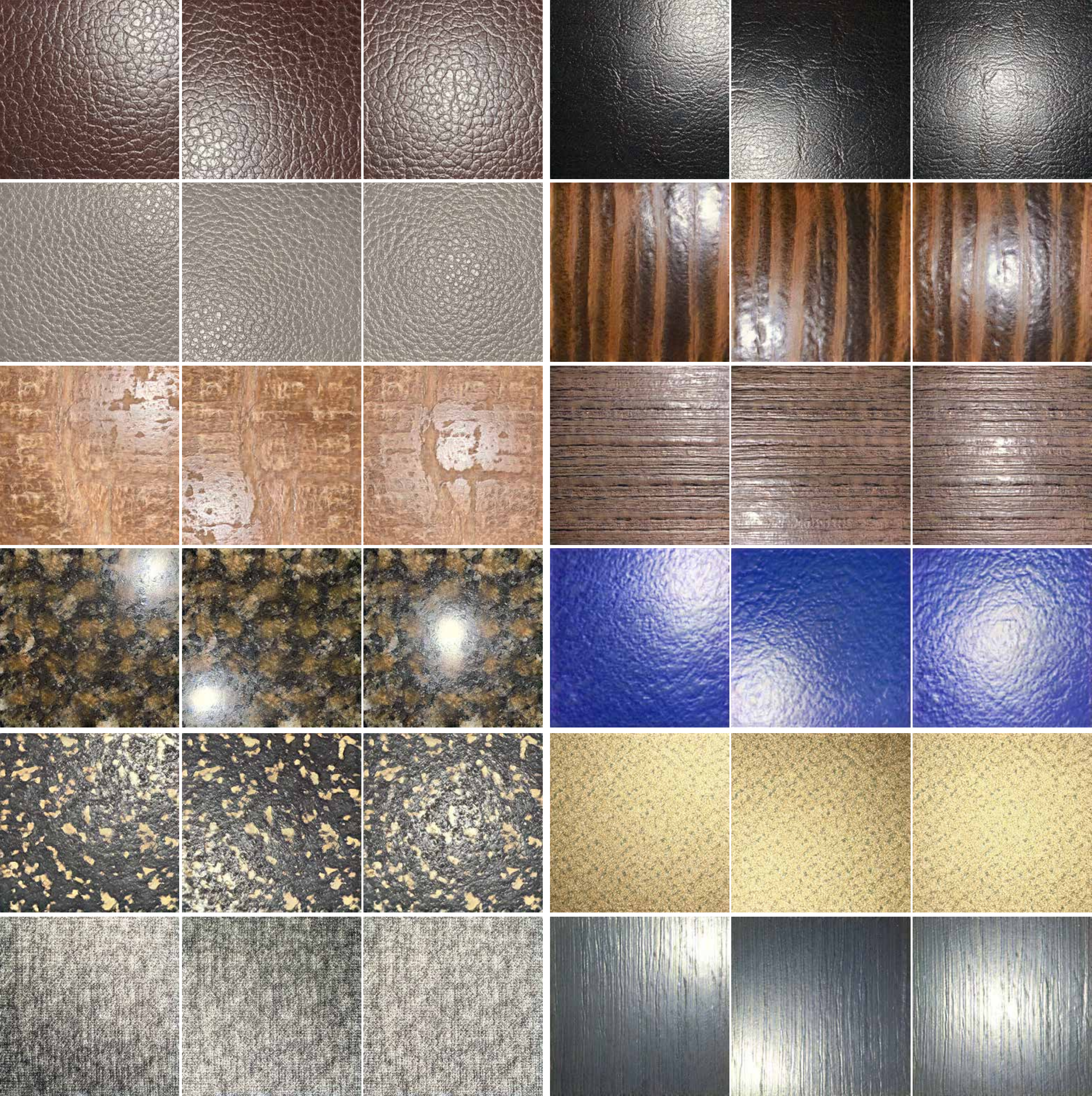}
\caption{Example outputs of our neural material generator: each triplet is a single neural material, $F$, relit under 3 different light locations. This model generates $256^2$ images and is trained on our small size (2.5k images) dataset.}
\vspace{-0.2in}

\label{fig:256_relight}
\end{figure}

\paragraph{Light falloff} An idealized omni-directional point light would have a falloff given by the inverse squared distance. However, the camera flash is usually not omni-directional and may have an additional falloff term, compounded with vignetting effects of the lens. Moreover, camera hardware variety, processing and tonemapping, and the ambient lighting in photos taken in a naturally lit environment, further complicate the computation of the exact falloff. In fact, the latter two effects can differ between images taken with the same camera, so no single falloff term will be accurate. \xz{Trying to calibrate the data acquisition more precisely would however conflict with our goal of in-the-wild capture.}

The imperfect falloff results in slight artifacts in the generated maps, where the estimator attempts to correct for the falloff by bending normals, and adjusting the intensity of other maps close to edges. We address this by finetuning the estimator with a different loss. In each iteration, we randomly shift the feature grid $F$ horizontally and vertically (with wraparound) before passing it to the maps estimator. We use a Gram matrix loss between the shifted and unshifted neural renders, which compares their overall style without pixel alignment. This removes falloff effects from the results, because the shifting withholds any information about which texels are close to the edges from the map estimator. This however comes at the cost of a small reduction in normal maps contrast. 

\paragraph{Summary} The material estimator $\decoder$ is designed to decode explicit reflectance properties from the learned implicit neural material representation trained on real materials. Our end-to-end system is very fast, as it combines forward inference of a generator with a U-net estimator. \xz{For $256^2$, $512^2$ and $1024^2$ models, sampling rates are 900, 200 and 60 samples/minute respectively, tested on a single RTX2080 GPU.} See Fig. \ref{fig:256_svbrdf} for the material maps estimated for the sampled materials from Fig. \ref{fig:256_relight}.
\section{Implementation}
\label{sec:impl}

In this section, we provide further details on the neural architectures, real dataset collection and processing, and training details.

Our neural material generator $\gen$ is built upon StyleGAN2~\cite{StyleGAN2} modified to output a feature tensor $F$ with the size of $H \times W \times 32$. We build a tileable version of $\gen$, using the strategy proposed in TileGen~\cite{zhou2022tilegen} to preserve tileable output. We do, however, observe that tileable generators are harder to train and have worse Fr\'echet inception distance (FID) \cite{heusel2017gans} and lower apparent visual quality than non-tileable generators; more research on high-quality tileable generation is needed.

Our relighting module $\mlp$ is an eight-layer MLP applied per-texel, the input of which is the concatenated feature vector from the corresponding texel of $F$ and a per-pixel light direction (a 3-dimensional unit vector) computed from the conditional light position $l$. Each inner layer of the MLP has 64 channels and we use LeakyReLU except at the last layer. The discriminator follows StyleGAN2 except that ours also takes per-pixel light directions concatenated to the RGB image input, for a total of 6-channel input. 

When fetching real images from the dataset, we crop $I_{crop}$ with resolution of $H$x$W$x$3$ from a real image $I_{real}$ with resolution of $2H$x$2W$x$3$. The crop coordinate is computed from conditional point light $l$ under the assumption that the camera is always collocated with point light source, and always at fixed height relative to the surface; even if not always true in practice, we can assume this without loss of generality. Only the ratio of camera distance and sample size matters in our results, and remains constant. We assume the camera distance to the sample equals the size of the visible sample area; we observe this is close to the true camera field-of-view of typical cell phones. 

For the maps estimator $\decoder$, we use a UNet~\cite{ronneberger2015u} with skip connections, which includes five downsampling and upsampling layers. This network takes $F$ as input and outputs certain channels $n$ of maps representing the material parameters (32$\rightarrow$32$\rightarrow$64 $\rightarrow$128$\rightarrow$256$\rightarrow$256$\rightarrow$256$\rightarrow$128$\rightarrow$64$\rightarrow$32$\rightarrow$$n$). Typically the material parameters include diffuse albedo, specular albedo, roughness, and normal or height (we have tested both; in the latter case the UNet \xz{produces a height map, which is then explicitly converted to normal map for rendering}). Our observations show that predicting height leads to better results; it could also be used for displacement mapping. Thus $n=8$ for this model. We further demonstrate that our model can fit custom BRDF models. For metallic paint, we use a different analytic model including two specular lobes each with its own albedo and roughness and one color albedo. In this model $n$ is set as 11. 




\paragraph{Training}

We follow the training strategy of StyleGAN2 to train $\gen$ and $\mlp$ with a learning rate of $0.0025$ with a batch size of 64. We train generation models at different image resolutions. We pick the $\gen$ and $\mlp$ with best FID during training. For $256^2$, $512^2$ and $1024^2$ model, we pick checkpoint at iteration 48k, 12k, and 11k respectively. We then freeze the pretrained $\gen$ and  $\mlp$, and use Adam ~\cite{kingma2014adam} with a learning rate (lr) of $1\times10^{-4}$ and batch size of 4 to train $\decoder$ for 60k iterations. Finally, we finetune the $\decoder$ by shifting $F$ during training for another 60k iterations with $lr = 5\times10^{-5}$. \xz{During this fine-tuning process, we use a Gram matrix loss and a random shifting strategy, further encouraging material maps to be invariant to spatial location.} We train $\gen$ and $\mlp$ on 8 V100 GPUs and train $\decoder$ on a single GPU.

\paragraph{Dataset collection}
In this section, we describe the rules we used to collect and prepare our real dataset. First, we attempt to center the flash light as much as possible ; we further run a simple highlight detector to help estimate the position of flash light in case of misalignment. Second, we minimize ambient lighting to ensure that the flash light dominates. Third, we strive to collect a diverse dataset covering multiple material categories that are easily available in everyday environments. 

Overall we present three datasets: a small \emph{Glossy} dataset containing 2.5k images of materials (limited  to  materials with low roughness and obvious specular reflection), another small dataset of 300 images of \emph{Car Paint materials}, which cannot be well reproduced with the standard GGX~\cite{Walter2007} model, and a larger ``in-the-wild'' dataset containing 9,000 images of materials covering many material categories (wood, stone, paint, leather, brick, metal, plastic, marble, fabric, paper, etc). We show examples of our ``in-the-wild'' dataset in Fig.~\ref{fig:real_dataset}. We combine all three datasets to build a larger 12,000-image \emph{Diverse} dataset. The \emph{Glossy} and \emph{Car Paint} datasets are used to train our $256^2$ and metallic paint models. The \emph{Diverse} dataset is used to train $512^2$ and $1024^2$ models.

To compute the light position, we start by computing the intensity map as the minimum of RGB channels per pixel. We gamma correct (2.2) the intensity map and compute its mean. We finally obtain the flash light position by computing the weighted average of all pixels with intensity value greater than this mean intensity. This method works for most examples, for the few examples where the light position is detected incorrectly, we manually set light position as the center of the image. \xz{The accuracy of light detection is illustrated visually in the supplementary materials.}
\section{Results}

In this section, we show the outputs of the relightable generator and material maps estimator for $256^2$, $512^2$ and $1024^2$ resolutions. We then demonstrate that our method can be extended to other material reflectance models by fitting a special smaller car paint dataset. We then compare PhotoMat with TileGen to show our results are more visually realistic and perform an user study for comparison. Finally, we demonstrate that PhotoMat can be extended to a tileable version either using HexTiling~\cite{mikkelsen2022practical} or with the tiling strategy proposed by TileGen.

\subsection{Neural Relit Materials and Analytical Material Maps}

\begin{figure*}
    \centering
    \includegraphics[width=1\linewidth]{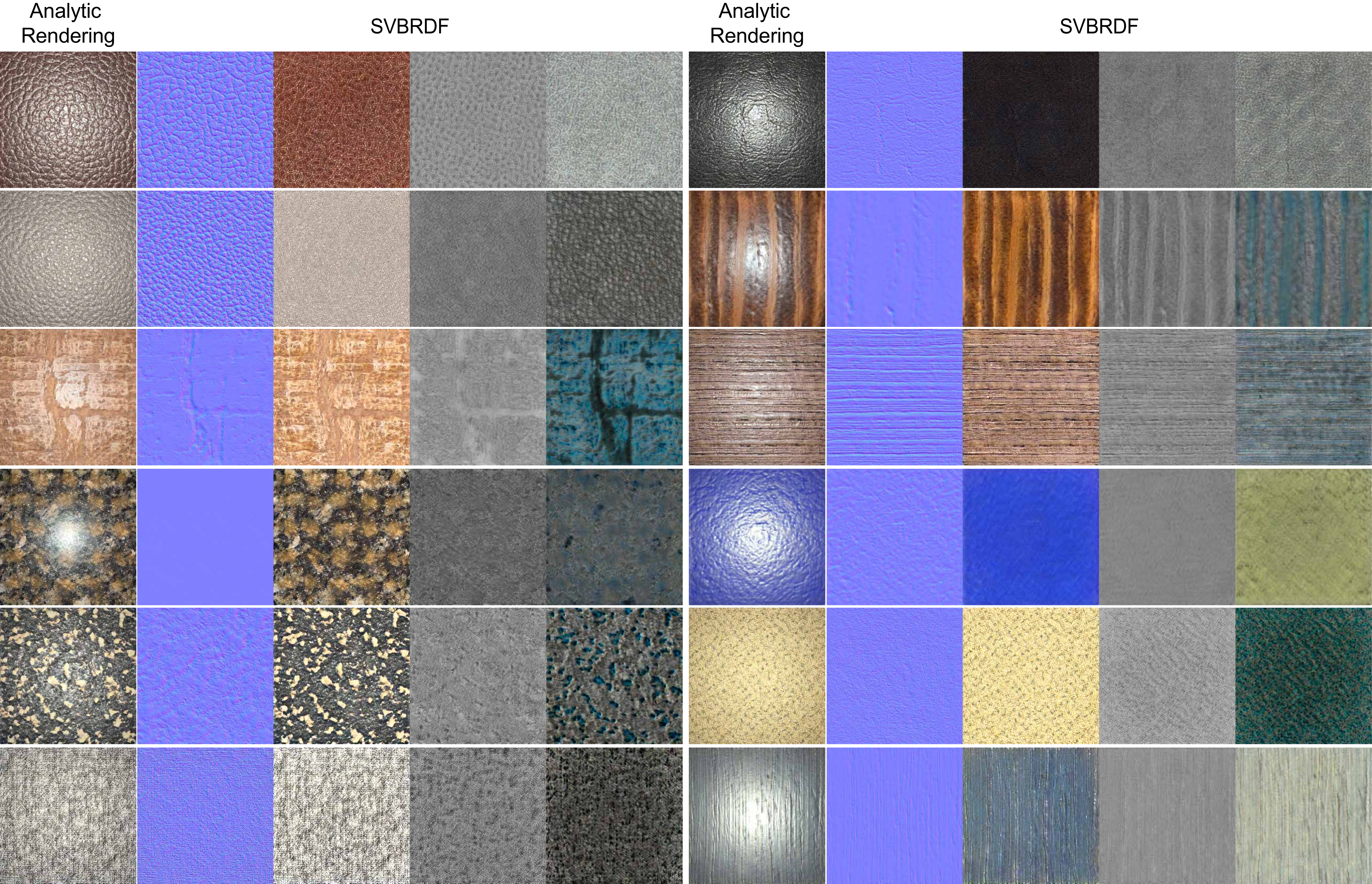}
\caption{The material parameter maps produced by our maps estimator $\decoder$ for the examples shown in Fig.~\ref{fig:256_relight}. The four maps are (left to right) normal, diffuse albedo, roughness and specular albedo. In practice, we estimate height and derive normal as its gradient. The occasional colored specular coefficient is due to imperfect white balance of the in-the-wild photos and can be easily corrected to grayscale if desired.}
\label{fig:256_svbrdf}
\end{figure*}

\paragraph{Low-resolution Results} We first show the results of a $256^2$ model trained on the small 2.5k \emph{Glossy} dataset. In Fig.~\ref{fig:256_relight}, we show the relit neural materials, where each triple share the same latent space $z$ but with different conditional point light sources. As shown in the figure, the relit neural materials cover diverse material categories, and relighting results are realistic and consistent under different illuminations. The material parameter maps (normal, diffuse albedo, roughness and specular albedo) produced by our maps estimator $\decoder$ and the analytic renderings are shown in Fig.~\ref{fig:256_svbrdf}. As can be seen, the analytic renderings using the estimated maps are comparable to the relighting results, demonstrating that the maps estimator can output material maps that reproduce the appearance of generated neural materials with minimal quality loss between relit neural materials and analytic renderings.

\begin{figure}
    \centering
    \includegraphics[width=0.9\linewidth]{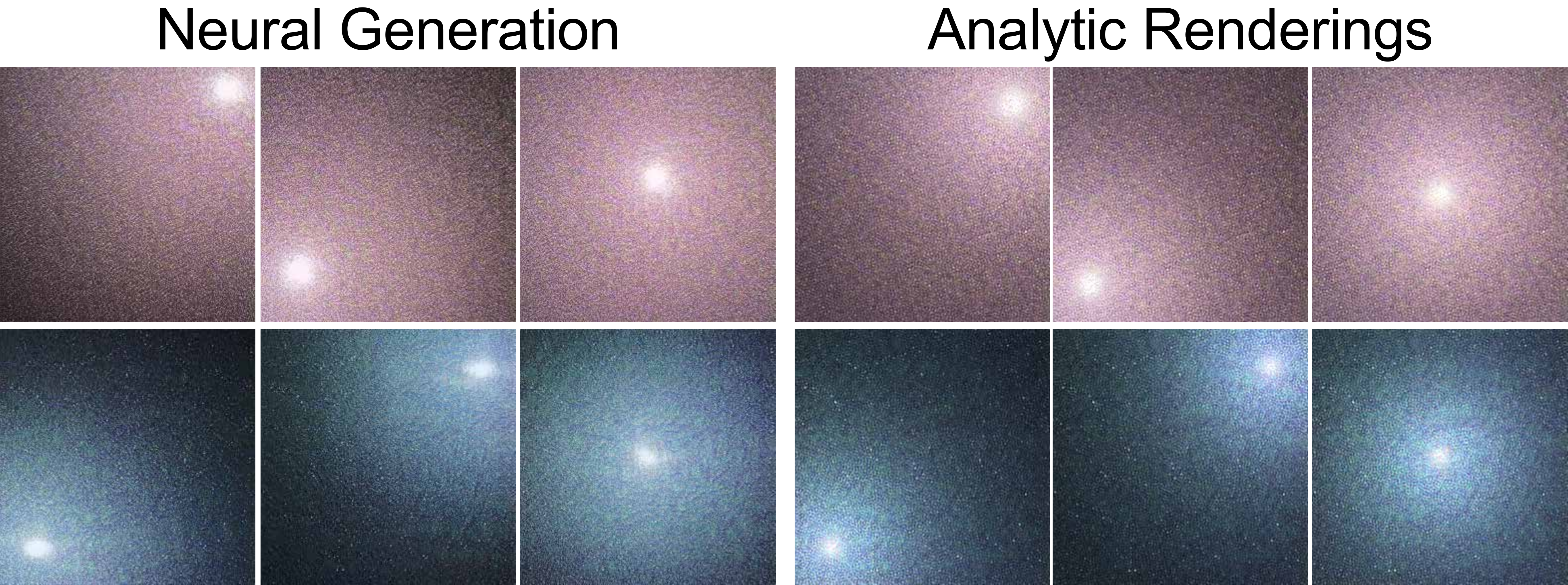}
\caption{We collect a small car paint dataset with challenging appearance (micro-flakes). For this material, we fit a two specular lobes analytic model, each with its own albedo and roughness, but no normals. We show that this approach generates realistic relit materials (Neural Generation) and the estimated analytic model matches them closely (Analytic Renderings).}
\label{fig:carpaint}
\end{figure}

\begin{figure}
    \centering
    \includegraphics[width=1\linewidth]{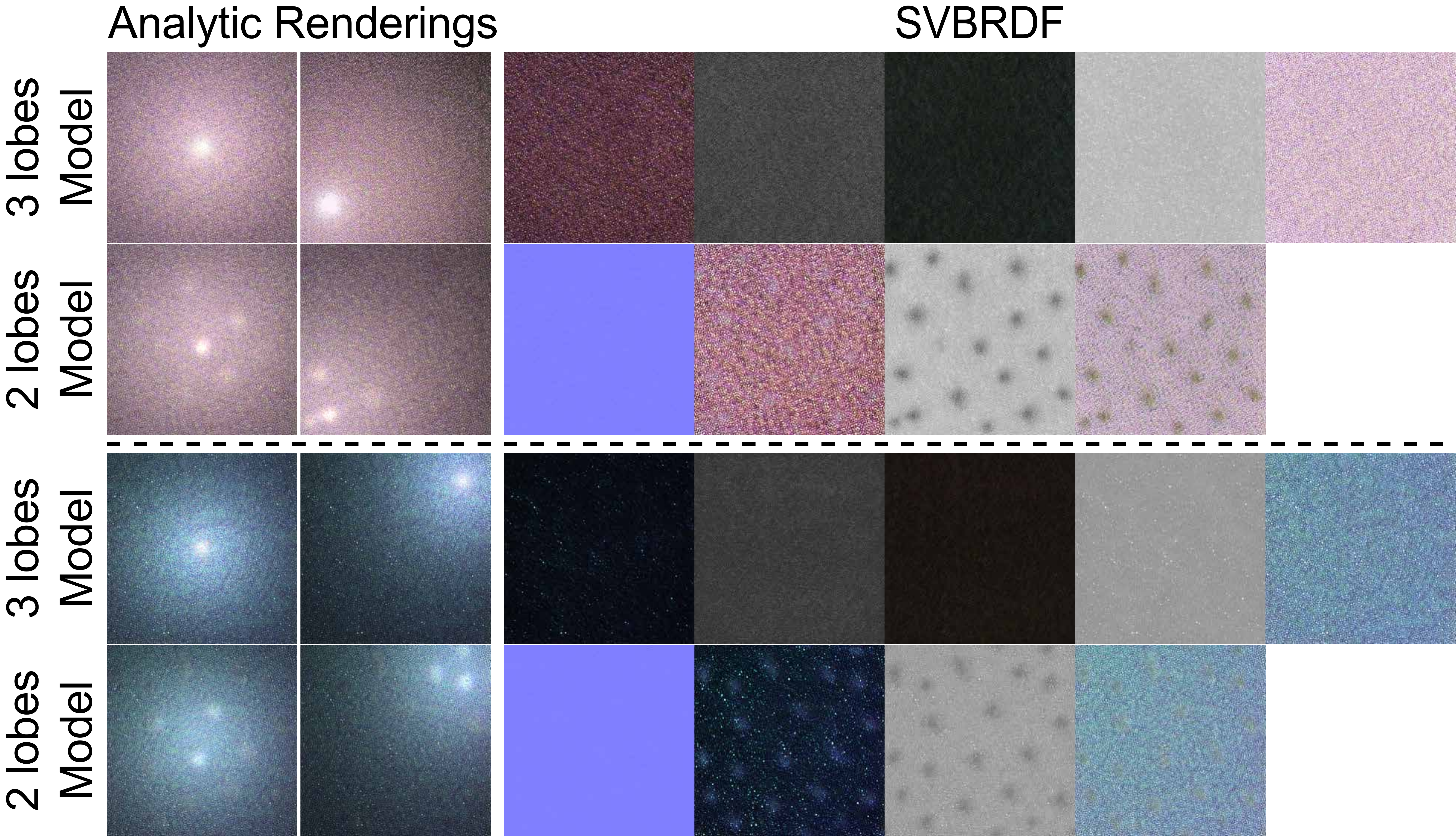}
\caption{Comparison of our multi-lobe analytic BRDF model (3-lobes model) for car paint with the standard BRDF model (2-lobes model) used for other materials. The 3-lobes BRDF includes one diffuse and two specular lobes, and each specular lobe has its own albedo and roughness. We see that the 2-lobes model cannot correctly fit the learned neural generator, producing multiple highlights and holes, but 3-lobes model succeeds.}
\vspace{-0.1in}
\label{fig:carpaint_compare}
\end{figure}

\paragraph{High-resolution Results}

\begin{figure*}
    \centering
    \includegraphics[width=1.0\linewidth]{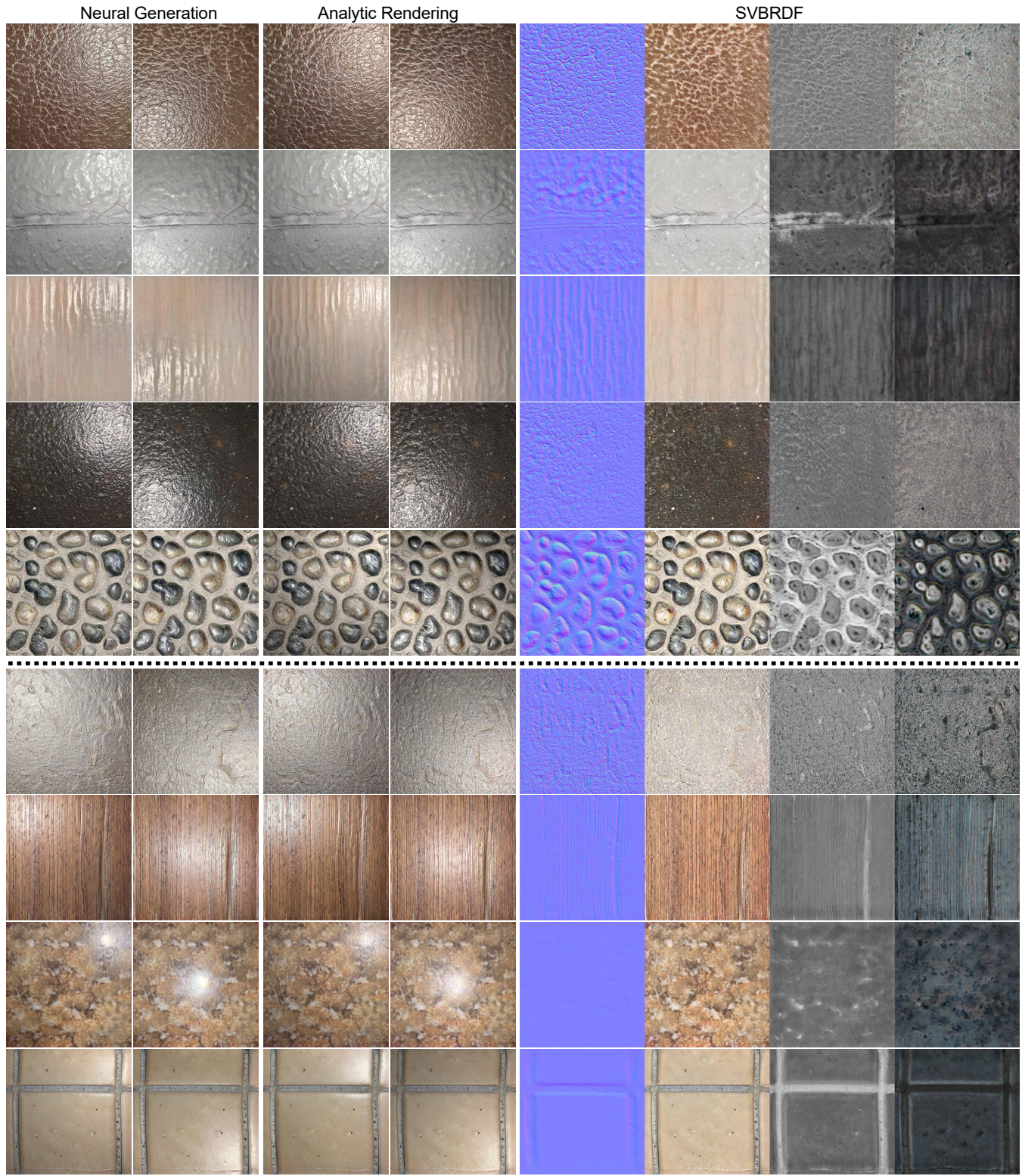}
\vspace{-0.25in}    
\caption{We train models of size $512^2$ (top) and $1024^2$ (bottom) on our largest \emph{Diverse} dataset. Here we show generated relightable materials (Neural Generation), corresponding renderings from estimated analytic BRDF parameters (Analytic Rendering) and the material maps (SVBRDF), showing normal, diffuse albedo, roughness and specular albedo.}
\label{fig:5121k_svbrdf}
\end{figure*}

We also train models of size $512^2$ and $1024^2$ on our larger \emph{Diverse} dataset, which covers more diverse material categories. In Fig. \ref{fig:5121k_svbrdf}, we show side-by-side comparisons of generated relit neural materials and the corresponding renderings from estimated analytic BRDF parameters, along with the estimated parameter maps. For this larger dataset, we use normal, diffuse albedo, roughness and specular albedo as material parameters. As is shown here, our higher resolution generator can also produce diverse and realistic relit materials, and the maps estimator can extract the reasonable corresponding materials maps even without supervision from ground truth material maps. \xz{We provide more visual results in supplementary materials.}


\subsection{Category-specific Model}

Our approach is not limited to a standard BRDF model and can be extended to any specialized BRDF model that is able to reproduce the appearance of a specific dataset. In Fig.~\ref{fig:carpaint}, we show results of model trained on the \emph{Car Paint} dataset. More specifically, we train a relightable generator on car paint photographs, and adjust the maps estimator to fit a coated BRDF model with one diffuse lobe and two specular lobes, each with its own albedo and roughness. As is shown in the figure, PhotoMat generates realistic relit neural materials (left) and the analytic renderings generated using the estimated SVBRDF matches them closely (right).

We further compare our special analytic BRDF model to the standard BRDF model used for other materials in Fig. \ref{fig:carpaint_compare}. We can observe that multiple holes are baked into the estimated maps for the standard model, since the standard BRDF model is not expressive enough to correctly fit the complex highlight falloff shape that our learned neural generator, trained on car paint dataset, captures. In comparison, our coated BRDF models produces high quality material maps that fit well to the relit neural materials. This illustrates that our approach is not tied to a specific BRDF model; neural generator easily extends to different material appearance and we can fit custom materials by training a corresponding maps estimator.

\subsection{Comparison against TileGen}

\begin{figure}
    \centering
    \includegraphics[width=1\linewidth]{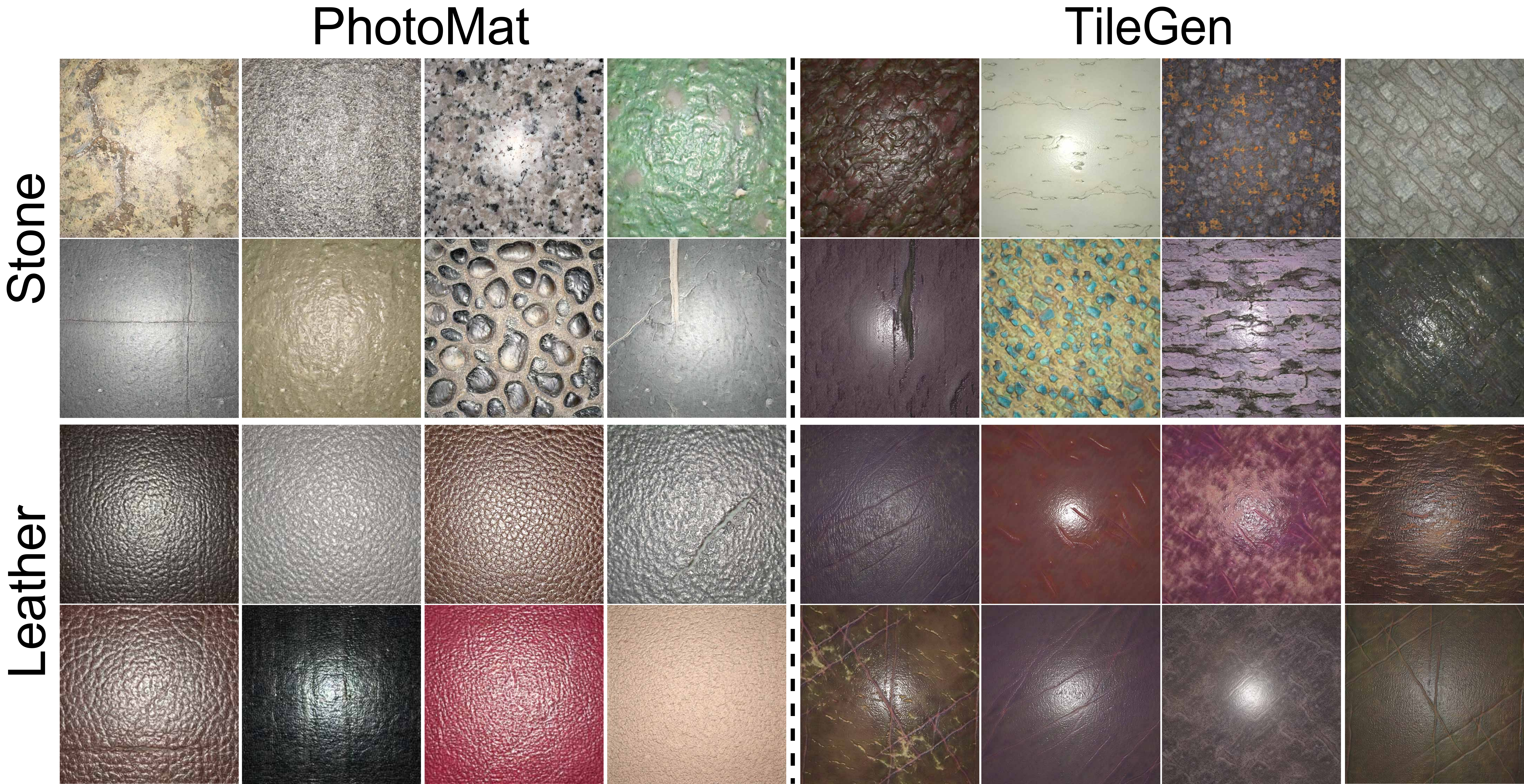}
\caption{For stone and leather, we compare a set of generated materials by our models against the models from TileGen, which are based on synthetic data. A larger version of this comparison can be found in the supplementary materials. We also conducted a user study, which confirms that our models generate more realistic materials overall.}
\vspace{-0.1in}
\label{fig:comparison_tilegen}
\end{figure}

We compare PhotoMat against TileGen \cite{zhou2022tilegen} visually and by conducting an user study. Note that TileGen is trained per category (tiles, leather, stone and metal), while our model is general. So we compare results for two material categories: stone and leather. For PhotoMat, we use analytic renderings using estimated analytic material maps and for TileGen we sampled results from leather and stone pretrained model. The visual comparison in Fig~\ref{fig:comparison_tilegen} obviously show that PhotoMat trained on real images produces more realistic appearance compared to TileGen trained on a synthetic dataset.  Please refer to figures in supplementary material for more visual comparison, where we add 30 materials per category for each method and show them side by side.

\paragraph{User study}
To evaluate this claim of realism of our method against that of TileGen, we conduct a user study with 30 participants with varied backgrounds in graphics. We show 20 randomly selected pairs of materials, 10 leather and 10 stones, from both our method and TileGen, in the form of flash lit renderings. We use a 2AFC method asking the users to select, for each stimulus, "Which image looks more like a photograph of a real commonly found surface, as opposed to an image synthetically designed by an artist?". The stimuli are randomly ordered and named. On average, participants find our method's results more realistic 79.2\% of the time. Further, out of 20 materials, 18 of ours are preferred and 1 shows equal preference. Out of 30 participants, 2 preferred TileGen results in the pairs (at 65\% and 55\%). The rest preferred ours (between 60\% and 100\%). This study confirms that training on real data does indeed result in more realistic generated SVBRDFs.

\subsection{Tileable and Extended Materials}

\begin{figure}
    \centering
    \includegraphics[width=1\linewidth]{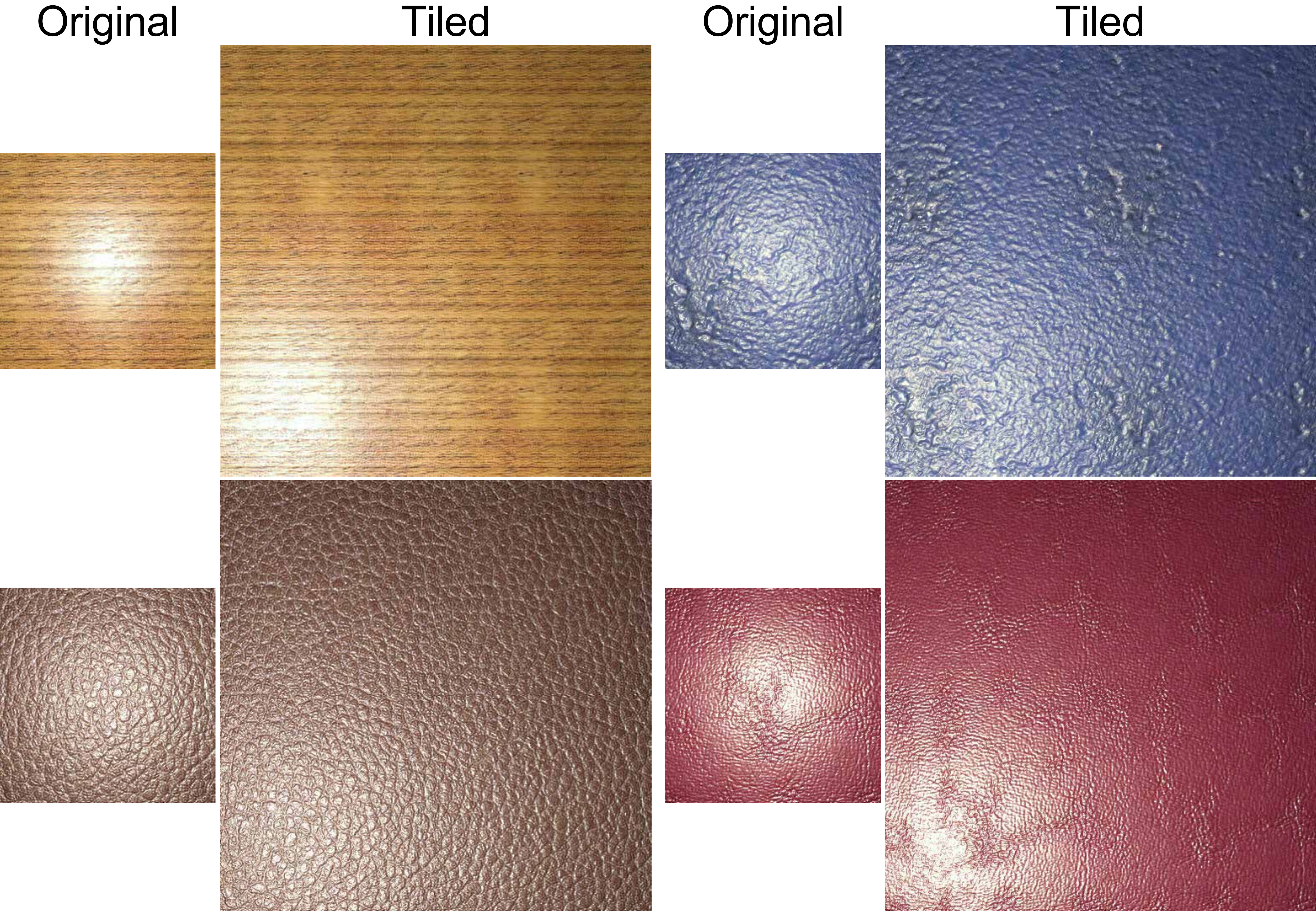}
\caption{We trained a $256^2$ model, tileable by construction, following the TileGen tiling strategy. This model produces natively tileable material maps; we show the renderings using original material maps alongside $2\times 2$ tiled versions.}
\label{fig:256_tile}
\end{figure}

\begin{figure}
    \centering
    \includegraphics[width=1\linewidth]{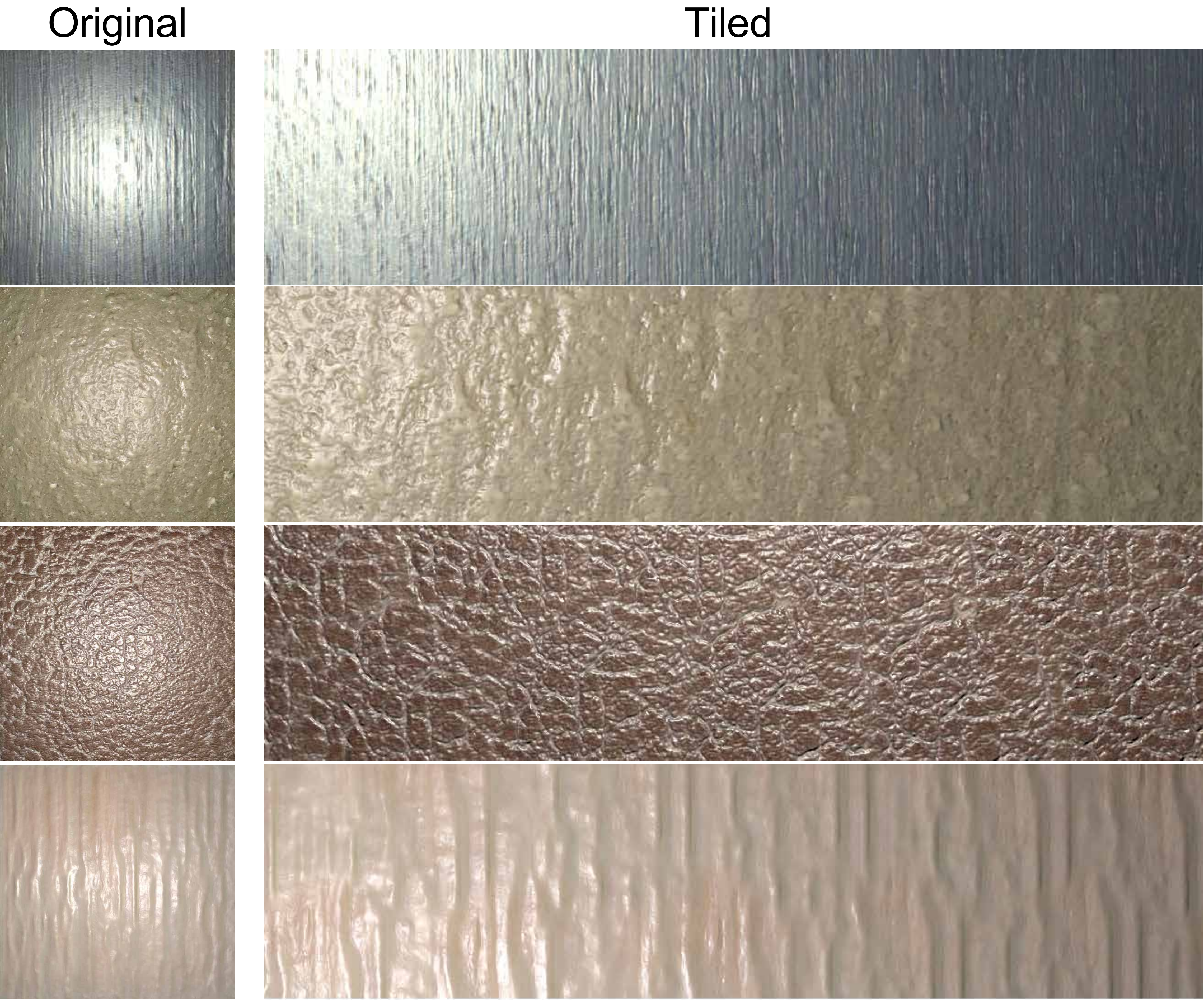}
\caption{ We apply the HexTile method \cite{mikkelsen2022practical} to tile our estimated material maps. The original rendered materials (left) are compared with tiled rendered materials (right), demonstrating that our estimated maps can be smoothly tiled and applied to the rendering pipeline. }
\label{fig:hextile}
\end{figure}

To extend the spatial extent covered by our generated materials, we demonstrate that PhotoMat can be extended to a tileable version by either following the strategy proposed by TileGen or using HexTile \cite{mikkelsen2022practical}. We train a tileable $256^2$ model following the unconditional TileGen's tiling technique. The original renderings and $2\times 2$ tiled renderings of tileable PhotoMat are shown in Fig.\ref{fig:256_tile}. Moreover, we also implement HexTile \cite{mikkelsen2022practical}, a previous method to smoothly blend hexagonal tiles extracted from the original material. The original renderings and tiled renderings are shown in Fig.\ref{fig:hextile}. Both strategies enable seamless renderings of PhotoMat materials, which can be passed to a rendering pipeline.




\section{Limitations}

\xz{Our dataset is naturally biased towards materials and scale ranges for which cell phone capture is convenient, and remains limited in scale. Scaling up our dataset size is likely to lead to the most dramatic improvements, much like with generative models in other domains~\cite{ramesh2022hierarchical}. We observed cases of mode collapse when training high-resolution and tileable versions of PhotoMat.  As mode collapse is a known drawback of GANs, different generative architectures such as diffusion models may be interesting to explore.
}
\vspace{-0.1in}

\section{Conclusion and Future Work}

In this paper, we show that it is possible to train a material generative model without any synthetic data, by using real flash-photographs dataset captured using a light-weight cell phone.
Since the desired reflectance maps of real materials (albedo, normal, roughness, etc.) are never directly observed, we solve the problem in two steps.

First, by simple cropping of the original photos with centered flash highlights, we obtain real training images with a variety of highlight locations. We train a GAN producing similar cropped flash images conditional on the flash highlight position. From the implicit neural material representation produced by the GAN, we train another neural network to extract per-texel analytic material parameters. We demonstrate a number of materials sampled from our generative model, applied in the context of a full rendered 3D scene with global illumination (see Fig.~\ref{fig:teaser} and supplemental material).

As it only requires a cellphone, our data collection can be easily scaled to large material datasets; we provide such a dataset of ~12,000 photos, and models trained on this dataset. While our trained models can be already used to generate material maps for practical use, we also believe that our approach opens the possibility of further extensions to different material models, other lighting conditions and larger real datasets. We believe this is a significant step in reducing the reliance of material research on synthetic data. This is especially important given the rapid progress in image generation demonstrated by diffusion models trained on extreme large-scale datasets. Extending these approaches to materials, for example to train conditional materials generators based on text or image prompts, would require approaches like ours to scale to these data requirements.

\begin{acks}

This project was funded in part by the NSF CAREER Award $\#2238193$ and a generous gift from Adobe.
\end{acks}

\bibliographystyle{ACM-Reference-Format}
\bibliography{references}

\end{document}